\documentclass[runningheads]{llncs}

\usepackage{eccv}

\usepackage{eccvabbrv}

\usepackage{graphicx}
\usepackage{booktabs}

\usepackage{adjustbox}
\usepackage{array}
\usepackage{makecell}
\usepackage{placeins}
\usepackage{listings}
\usepackage{xcolor}
\usepackage{pifont}%
\DeclareMathOperator*{\argmax}{arg\,max}

\definecolor{codegreen}{rgb}{0,0.6,0}
\definecolor{codegray}{rgb}{0.5,0.5,0.5}
\definecolor{codepurple}{rgb}{0.58,0,0.82}
\definecolor{backcolour}{rgb}{0.95,0.95,0.92}

\newcommand{\cmark}{\color{ForestGreen}\ding{51}}%
\newcommand{\xmark}{\color{red}\ding{55}}%

\lstdefinestyle{mystyle}{
    backgroundcolor=\color{backcolour},   
    commentstyle=\color{codegreen},
    keywordstyle=\color{magenta},
    numberstyle=\tiny\color{codegray},
    stringstyle=\color{codepurple},
    basicstyle=\ttfamily\footnotesize,
    breakatwhitespace=false,         
    breaklines=true,                 
    captionpos=b,                    
    keepspaces=true,                 
    numbers=left,                    
    numbersep=5pt,                  
    showspaces=false,                
    showstringspaces=false,
    showtabs=false,                  
    tabsize=2
}

\usepackage[accsupp]{axessibility}  %

\usepackage{hyperref}

\usepackage{orcidlink}

\newcolumntype{R}[2]{%
    >{\adjustbox{angle=#1,lap=\width-(#2)}\bgroup}%
    l%
    <{\egroup}%
}
\newcommand*\rot{\multicolumn{1}{R{70}{1em}}}%

\begin{document}

\title{CLAMP: Contrastive Language Model Prompt-tuning  }

\titlerunning{CLAMP: Contrastive Language Model Prompt-tuning  }

\author{Piotr Teterwak\inst{1} \and
Ximeng Sun\inst{1} \and
Bryan A. Plummer\inst{1} \and
Kate Saenko\inst{1} \and
Ser-Nam Lim\inst{2}}

\authorrunning{P.~Teterwak et al.}

\institute{Boston University \\ \email{piotrt@bu.edu} \and
University of Central Florida}

\maketitle

\begin{abstract}
Large language models (LLMs) have emerged as powerful general-purpose interfaces for many machine learning problems. Recent work has adapted LLMs to visual tasks like image captioning, visual question answering, and visual chat, using a relatively small amount of instruction-tuning data. Surprisingly, as we show in this paper, the resulting multimodal-LLMs (mLLMs) are poor at classifying an image into a set of categories, and far worse than specialized models like CLIP. We explore whether modern LLMs can be adapted to perform this fundamental visual task, and propose an approach for paremeter-efficient fine-tuning of LLMs using the same contrastive image-caption matching objective as CLIP. Our results show that LLMs can, indeed, achieve good zero-shot image classification performance. Our approach beats state-of-the-art mLLMs by 13\% and slightly outperforms contrastive learning with a custom text model (known as LiT), while also retaining the LLM's generative abilities. LLM initialization appears to particularly help classification in domains under-represented in the visual pre-training data.

\end{abstract}
    
\section{Introduction}
\label{sec:intro}

\begin{figure}[t!]
  \centering
   \includegraphics[width=\linewidth]{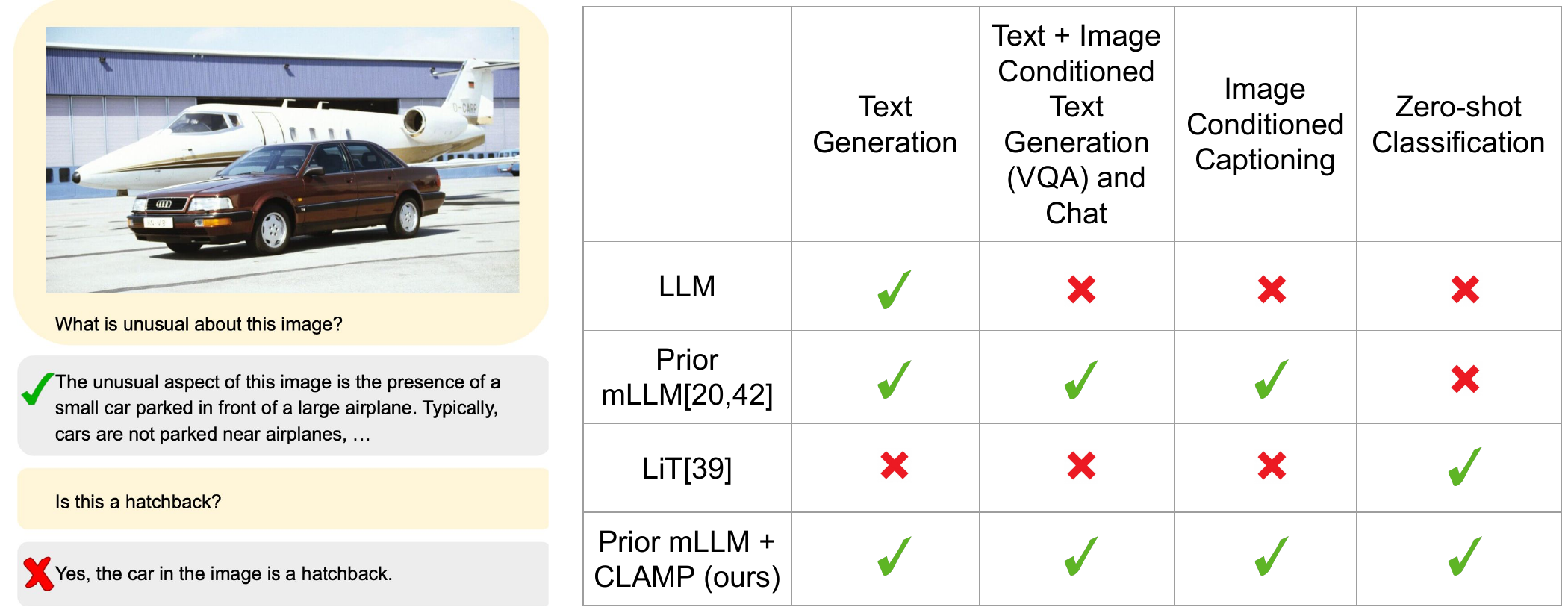}

   \caption{State-of-the-art multimodal LLMs excel at generative visual tasks like answering questions that involve common sense, but underperform on standard image classification tasks like predicting the car type (generated using LLaVA1.5~\cite{liu2023visual}). On a suite of 24 zero-shot image classification datasets they underperform SOTA zero-shot classification models like CLIP \cite{radford2021learning} by 13\%.  
   In this paper, we present CLAMP, an approach to add  classification abilities to a base LLM. This extends an LLM's visual reasoning ability to include visual discrimination, a fundamental computer vision task that true foundation models need to have. Putting together prior mLLM adapter modules and CLAMP, LLM's are now able to generate text, answer visually-grounded questions, chat interactively, and do zero-shot object classification. }
   \label{fig:one}
\end{figure}

In the past several years the capabilities of Large Language Models (LLMs) have progressed from interesting to extremely useful. %
Recent work has adapted LLMs to multimodal inputs such as images, 
by learning a mapping from a frozen visual backbone to a continuous token used as  input to the LLM. Such multimodal LLMs (mLLMs) include  LLaVA~\cite{liu2023visual}, MiniGPT~\cite{zhu2023minigpt},  Instruct-BLIP~\cite{instructblip},  and GPT-4V~\cite{openai2023gpt} and have impressive image captioning and question-answering capabilities. For example, they can use the LLM's world knowledge to effectively answer common sense questions, such as why a photo is unusual (Fig.~\ref{fig:one} left).

One might assume that the LLM's world knowledge would similarly help in classifying objects into categories, such as what type of car is in the image (Fig.~\ref{fig:one}). After all, an LLM can provide a detailed description of what different cars look like. In this paper we ask, \textbf{can  multimodal LLMs also perform zero-shot classification tasks?}  For instance, given a list of vehicle types and an image, can an mLLM identify which type of vehicle is present? We find that the answer is, surprisingly, not well. For example, state-of-the-art mLLMs get under 10\% top-1 classification accuracy on Stanford Cars \cite{KrauseStarkDengFei-Fei_3DRR2013}, while CLIP achieves over 80\% (Tab. \ref{table:results}). This is despite the fact that the best classification algorithm for current mLLMs computes the likelihood for \textbf{each} image-label pair, inefficiently requiring many passes through the LLM (see Section \ref{sec:coverage}, Runtime). Poor classification performance of mLLMs is counter-intuitive;  LLMs have been proven to encode extensive knowledge about the world. %
Furthermore, visual classification is a fundamental computer vision task, and is a capability we definitely want a foundation model to have. 

We hypothesize that this weakness is due not to a lack of capability of the LLM but to improper training,
and that the generative captioning and instruction-tuning objectives do not enable effective classification. To build some intuition for this, consider that LLMs contain a powerful language prior which may improve visual-question answering even without image features; for example, the correct answer may simply have higher-likelihood in the language model, as illustrated by Visual GPT-Score \cite{lin2023visualgptscore}. In that work, it was shown that SoTA performance can be achieved on several vision-language benchmarks by ignoring the context image completely and just computing the likelihood of the text sequences using the language model. 

\begin{figure}[t]
  \centering
   \includegraphics[width=\linewidth]{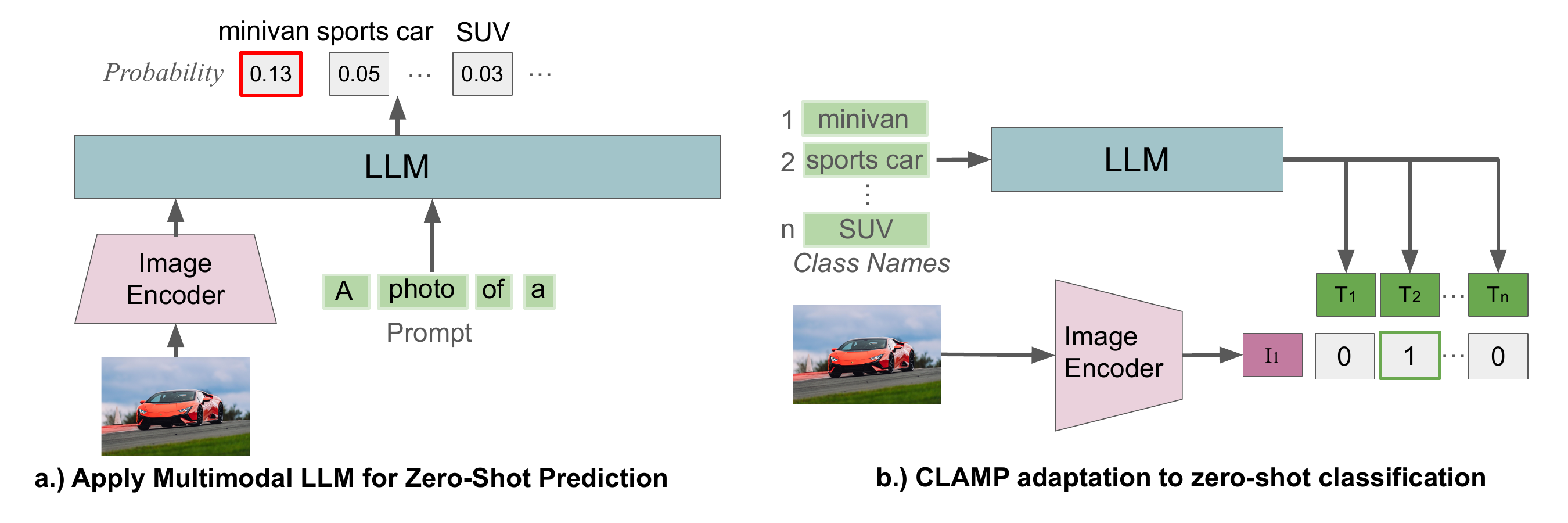}
   \caption{\textbf{Adapting LLMs for image classification:} a) Applying prior multimodal LLMs such as LLaVA~\cite{liu2023visual} and MiniGPT~\cite{zhu2023minigpt} to classification by computing the GPTScore \cite{lin2023visualgptscore} has poor accuracy; b) Our approach CLAMP achieves high accuracy by lightly fine-tuning the LLM with a contrastive image-caption objective.} \label{fig:ours}
\end{figure}

Given our hypothesis that generative training is not sufficient for discriminative image tasks, but that LLMs still encode  knowledge useful for those tasks, we propose to replace the text encoder of a vision-language pretrained  model like CLIP  \cite{radford2021learning}) with an LLM. We call our method \textit{Contrastive LAnguage Model Prompt-tuning (CLAMP)} (Fig.~\ref{fig:ours}.)  
 We find that by updating a minority of the LLM parameters, our method can approach the performance of zero-shot CLIP trained on much larger scale data while outperforming a text-encoder trained from scratch on the same data (LiT~\cite{zhai2022lit}). This finding highlights the benefits of using Large Language Models; LiT earlier found that initializing to pre-trained (but smaller scale) LMs is not helpful for zero-shot classification in English. We instead find that carefully finetuning modern, large language models is helpful relative to training a text-encoder from scratch, especially on evaluation datasets which have poor coverage in data used for contrastive training.  Surprisingly, we're \textit{also} able to mostly retain the generative capabilities of the language model, pointing towards universal generative and discriminative models. Finally, because we use parameter-efficient finetuning methods for our training and the same base LLM  as current instruction-tuned mLLMs, our method makes it possible to swap CLAMP's finetuned parameters with those of mLLMs like LLaVA. As we illustrate in Fig.~\ref{fig:one}, this enables captioning, VQA, chat and classification with the same foundational LLM.    %

In summary, our contributions are as the follows:
\vspace{-.1cm}
\begin{itemize}
    \item We show that SOTA multimodal LLMs are neither accurate nor efficient at zero-shot image classification. 
    \item We train a set of adapter modules with a contrastive loss, which enable an LLM to be used for visual discrimination tasks by replacing a text encoder in a contrastive vision-language model. 
    \item We show that the LLM initialization in CLAMP allows for improved performance in domains under-represented in the pre-training data, and that our CLAMP retains generative abilities despite our fine-tuning. 
\end{itemize}

\section{Related Work}

\noindent\textbf{Large Language Models.} Large Language Models are transformers \cite{vaswani2017attention}, scaled to billions of parameters, trained generatively on trillions of tokens. LLMs are commonly trained with next-token prediction.  The  success of LLM-based chatbots such as ChatGPT has inspired a great number of efforts to create both general \cite{touvron2023llama,openai2023gpt,zheng2023judging,chung2022scaling} and domain specific \cite{singhal2023large, wu2023bloomberggpt} language models. Instruction-tuning additionally fine-tunes LLMs with (instruction, output) pairs, which makes LLMs more useful for specific tasks. In this work, we explore adapting an LLM to image classification.

\noindent\textbf{Multimodal Large Language Models.}  Until recently, large LMs have operated exclusively in the text domain. However, a plethora of models such as LLaVA~\cite{liu2023visual}, GPT-4V~\cite{openai2023gpt}, MiniGPT~\cite{zhu2023minigpt}, Flamingo~\cite{alayrac2022flamingo}, and BLIP~\cite{instructblip} now can take images as input. These models typically use a pre-trained vision encoder and a small set of learnable parameters to map from vision encoder output to LLM input. They are typically trained with a captioning loss, and can additionally be tuned with an instruction-tuning objective. The base LLM can be frozen \cite{zhu2023minigpt} or updated \cite{touvron2023llama}. Multimodal LLMs work well for tasks such as visual question answering or captioning. However, we show that they are generally ineffective for zero-shot classification. But classification is a fundamental computer vision task, and one for which the world knowledge of LLMs is intuitively very useful. We show that with a Parameter-efficient fine-tuning, LLMs can perform discriminative visual reasoning.

\noindent\textbf{Regularized Fine-tuning.} Also known as Parameter-Efficient Fine Tuning (PEFT), regularized fine-tuning aims to update only a subset of the parameters in a pre-trained model. This allows one to keep around only a small number of parameters for each specialized task a model is fine-tuned for, to use less memory in the training process, and to avoid unlearning knowledge present in the pre-trained model. Common approaches include low-rank updates of existing matrices (LoRA \cite{hu2021lora}), prompt and prefix tuning \cite{lester2021power,liu2021p,lee2023rpo}, and inserting additional trainable layers into the middle of the network (Adapters \cite{houlsby2019parameter}). In our work, we use prompt tuning and LoRA to adapt the LLM to the contrastive text-image task.

\noindent\textbf{Contrastive Training for Vision-Language Models.} Contrastive training has become a popular training technique. It aims to align positive pairs in embedding space, while pushing negative pairs apart. Therefore, this type of method strongly depends on how positive and negative pairs are formed. %
Vision-Language models (\eg CLIP \cite{radford2021learning}) use a text-encoder and contrast against text captions \cite{radford2021learning,zhai2022lit,zhai2023sigmoid}, therefore learning a joint image-text embedding.  This joint embedding space enables Vision-Language models to classification in the zero-shot setting, without training data with explicit labels. In this work, we add a contrastive objective to fine-tune an LLM, and therefore enable the LLM to function as an effective zero-shot visual classifier.

\begin{figure}[t]
  \centering
   \includegraphics[width=\linewidth]{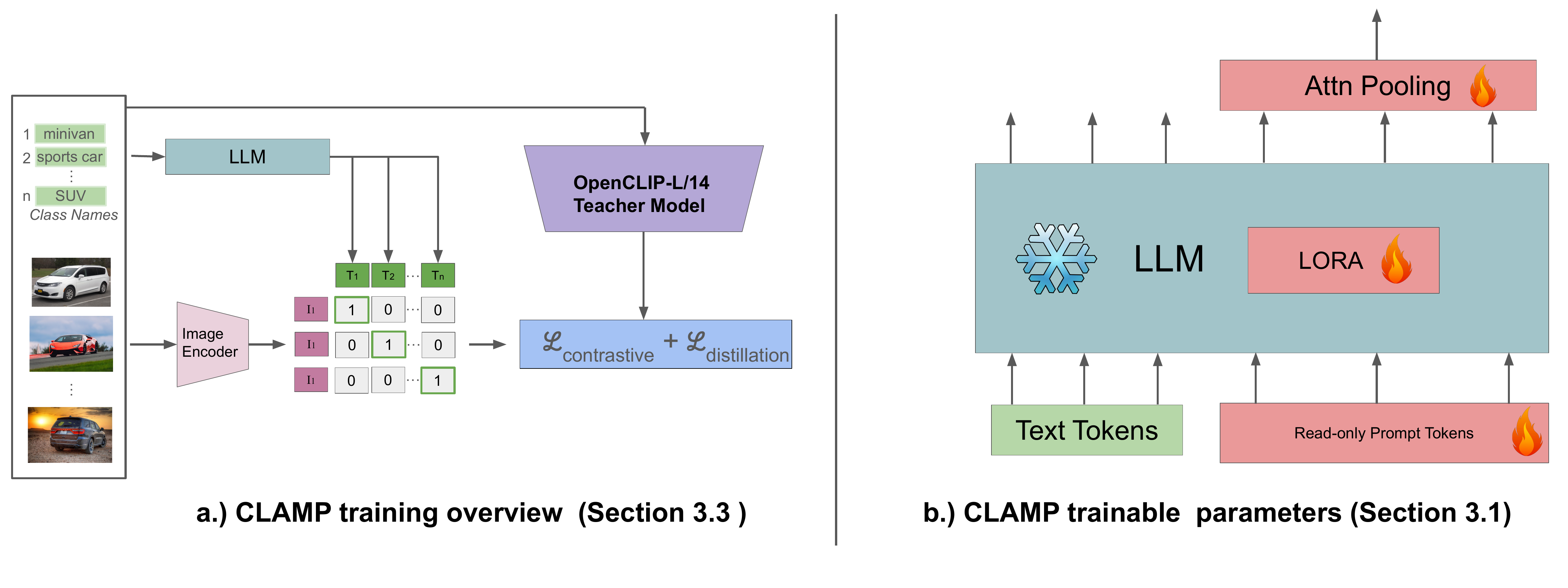}
   \caption{\textbf{Training CLAMP:} a) The overall training loss of CLAMP. CLAMP is trained with a CLIP loss together with a distillation loss. b.) An overview of trainable parameters. We combine Read-only Prompt Optimization, LORA, and Attention Pooling.  } \label{fig:ours_train}
\end{figure}

\section{CLAMP: Contrastive LAnguage Model Prompting }

We propose \textit{Contrastive LAnguage Model Prompt-tuning (CLAMP)}, a method to adapt Large Language Models to classification via prompting and contrastive learning.
  Our approach is inspired by state-of-the-art vision-language models trained with the contrastive objective, \ie, CLIP. The objective encourages a strong alignment between text  and image representations in a joint space, enabling effective-zero shot classification. We  propose to align LLM representations with visual representations using a contrastive loss. However, the output embedding space of a frozen LLM  may not be naively easy to align with a visual encoder for a discriminative task, because the LLM was trained for next-token prediction instead of cross-modal alignment. Instead of using a fully frozen LLM, CLAMP uses PEFT (Parameter-Efficient Fine-Tuning) to lightly update the model to align it with CLIP visual encoders.

CLAMP accomplishes this in three main ways (Section \ref{method:attn_pool}). First, it learns a pooling over the LLM's output representations to aggregate information over the entire sequence. We implement this with attention-pooling, with a query token that learns to aggregate the output representations of an encoded caption. However, output pooling is insufficient to get the LLM to produce useful information that is embedded in the remaining network layers.  To overcome this, CLAMP adds learned read-only prompts \cite{lee2023rpo}, which serve as special-purpose discriminative tokens in a model pre-trained for generation. This enables CLAMP to extract information from intermediate network layers, to align LLM embeddings with visual embeddings. Finally, we find that slight modification of the parameters improves zero-shot performance further, accomplished with LoRA, which learns low-rank delta updates of certain layers.

Section~\ref{method:wrapper_prompts} details the text prompt used in training and Section~\ref{method:training} describes the training objective, which includes  distillation of a stronger CLIP model into our trained model, inspired by Dime-FM \cite{Sun_2023_ICCV}. 

\subsection{Architecture}
Our goal is to align an LLM with a pre-trained visual encoder contrastively (see Fig. \ref{fig:ours_train}). In our experiments, we use the Vicuna-1.5 LLM\cite{vicuna2023}, which is a LLaMA-2 based instruction-tuned model. In order to avoid overfitting, we only add a small number of trainable modules including Output Attention Pooling at the output, Read-Only Prompts at the input, and LoRA in the intermediate layers. Similar to many mLLMs, we freeze the visual encoder.

\subsubsection{Read-only Prompts}
\label{method:rpo}

Frozen LLMs are generative models, therefore in order to add discriminative capabilities we need to modify the model. However, doing so without harming the good representations learned during pre-training requires care, and one method used to do so is prompt tuning. We append learnable prompts to the end of the text sequence, which function as special-purpose tokens for discriminative tasks: 
\begin{equation}
x = [x_{\text{text}} ; p_{\text{learnable}}]
\end{equation}
where $x$ is the input sequence to the LLM, $x_\text{text}$ is the tokenized language and $p_{\text{learnable}}$ are the learnable prompts. 
The LLM we use has causal attention, meaning each position's representation can only attend to tokens previously seen in the sequence. When attention is bi-directional, the intermediate representations of $x_\text{text}$ can shift with changes in $p_{\text{learnable}}$.  With causal attention, however, $p_{\text{learnable}}$ can change without updating intermediate representations of $x_\text{text}$. Such a technique is known as read-only prompting \cite{lee2023rpo}, and it is shown to improve generalization of prompt-tuning of contrastive vision-language models to unseen classes, due to an effect of minimizing overfitting. To produce the text embedding, we perform output attention pooling over the output representations of the learnable prompt tokens.

\subsubsection{Output Attention Pooling}

\label{method:attn_pool}

Transformer networks take as input variable-length sequences, and output sequences of that same length. However, the joint image-text embedding space of vision-language models requires a representation of a fixed length. Therefore, some kind of pooling of tokens is needed. CLIP \cite{radford2021learning} accomplishes this by simply using the last token in the sequence as the representation of the text.  We instead use the read-only prompts described in the previous section. However, since we use many of those prompts, we still need to pool them to a fixed length. We choose to leverage a form of pooling that uses  the multi-headed attention mechanism, where $W_Q$, $W_K$, $W_O$, and $W_V$ are parameter matrices. 

\begin{equation}
 \text{Attention}(Q, K, V) = \text{Concat}(\text{head}_1, \ldots, \text{head}_h)W_O \\
\end{equation}
where
\begin{equation}
\text{head}_i = \text{Attention}(QW_{Qi}, KW_{Ki}, VW_{Vi}) \\
\end{equation}
and 
\begin{equation}
    \text{Attention}(Q, K, V) = \text{Softmax}\left(\frac{QK^T}{\sqrt{d_k}}\right)V
\end{equation}

In our attention pooling mechanism, we learn the query Q and the K and V are simply output tokens of the LLM. In this way, we can learn a per-sample linear combination of LLM outputs.

\subsubsection{LoRA}
\label{method:lora}

While Attention Pooling and Read-Only Pooling are important architectural components in extracting discriminative features from text sequences, discriminative performance can further be improved if the network parameters can be lightly updated. One option is full-finetuning of the network parameters. This, however, has two large drawbacks. First, it has heavy memory requirements. In particular, for the large-batch training necessary for effective contrastive learning, full fine-tuning requires the storage of intermediate activation. Secondly, we go back to the motivation of not catastrophically forgetting information present in the LLM in our contrastive training. Therefore, we leverage LoRA, which learns low rank updates to pretrained models:

\begin{equation}
    W = W_0+ BA^T
\end{equation}

\noindent where $W$ are the updated weights in $\mathbb{R}^{d \times d}$, and $B$ and $A$ are in $\mathbb{R}^{d \times k}$.

As is common, we only update multi-headed attention layers with our LoRA updates. See the Supplementary for exact LoRA configuration. %

\subsection{Wrapper prompts}
\label{method:wrapper_prompts}

Vicuna is trained with a very specific prompt template: \vspace{5pt}\\
``\textit{A chat between a curious user and an artificial intelligence assistant. The assistant gives helpful, detailed, and polite answers to the user's questions. \\
USER: Hello! \\
ASSISTANT: Hello!".} \vspace{5pt}

In order to follow this format of USER followed by ASSISTANT, we  use the following prompt template for our training: \vspace{5pt}\\
\textit{``USER: A photo of a \{\}. A \{\} has the following visual attributes.\\
ASSISTANT: 1.''}\vspace{5pt}\\
where the \{\} is replaced with the caption of a particular sample. At test time, \{\} is replaced with the class name.   Then, this text becomes $x_{\text{text}}$ and the final input to the CLAMP LLM is $x = [x_{\text{text}} ; p_{\text{learnable}}]$.  The \textit{ASSISTANT: 1}, would prompt the model to generate a response, if it were being using auto-regressively.  Auto-regressive decoding of this prompt would result in a description of what the image with the caption may look like. Although auto-regressive decoding in the training loop is prohibitively expensive; Xiao~\etal \cite{xiao2023efficient} show that much information about the decoded sequence is present in the initial tokens, and therefore prompting the LLM to describe the caption is a reasonable way to surface visual information.

\subsection{Contrastive Training of LLMs}\label{method:training}

So far, we have discussed the architectural components of our method. We now turn our attention to the training methodology. Following CLIP \cite{radford2021learning}, we use image-text and text-image contrastive losses:
\begin{equation}  
\mathcal{L}_{img2txt}	=  - \sum_{ i }  
\log \frac{ \exp(\tau v^T_i t_{i})  }{\sum_{ j }  \exp(\tau v_i^Tt_{j})  }
\end{equation}
\begin{equation}  
\mathcal{L}_{txt2img}	=  - \sum_{ i }  
\log \frac{ \exp(\tau t^T_i v_{i})  }{\sum_{ j }  \exp(\tau t_i^Tv_{j})  }
\end{equation}
\begin{equation}
    \mathcal{L}_{contrastive} = \mathcal{L}_{img2txt} + \mathcal{L}_{txt2img} 
\end{equation}
where $t$ is a text representation, $v$ is a visual representation, and $i$ and $j$ index the batch. $\tau$ is  a learned temperature parameter.

In addition, we add a distillation loss term to the training, similar to Dime-FM \cite{Sun_2023_ICCV}. Following \cite{Sun_2023_ICCV}, we first compute teacher similarity matrix $S_t$ and student similarity matrix $S_s$ using the vision and text encoders over samples in the batch, as in the contrastive loss. Defining row-wise indexing variable as $i$ and column-wise indexing variable as $j$, the distillation loss is defined as:
\begin{equation}
    \small
    \mathcal{L}_{dist} = \sum_{i} KL( \tau_d S_t || \tau_d S_s) + \sum_{j} KL( \tau_d S_t^T || \tau_d S_j^T)
\end{equation}
 Finally, our loss becomes:
\begin{equation}
    \small
    \mathcal{L}_{tot} = \mathcal{L}_{contrastive} + \mathcal{L}_{dist} 
\end{equation}

During inference time, we simply take the argmax over classes of the dot  product of an input image features $x$ and the CLAMP representations of class labels $t_c$ to produce predicted class label $y_{pred}$.

\begin{equation}
    \small
    y_{pred} =\argmax_c{xt_c^T} 
\end{equation}

\section{Experiments}

We evaluate our models on zero-shot classification tasks in order to test the discriminative ability of our adapted large language models. We compare to two sets of models; \textbf{a.)} To existing multimodal Large Language Models  LLaVA \cite{liu2023visual} and Mini-GPT \cite{zhu2023minigpt}, to demonstrate how naive usage of existing mLLMs does not meet the expectations of modern zero-shot classifiers, and \textbf{b.)} To a contrastive vision-language model where we retrain the parameters of the text encoder, also known as LiT \cite{zhai2022lit}. We select a diverse set of 24 datasets to evaluate zero-shot performance, including fine-grained datasets such as Stanford Cars \cite{KrauseStarkDengFei-Fei_3DRR2013} and Aircraft \cite{maji2013fine}, natural but challenging datasets such as EuroSAT\cite{helber2017eurosat} and ImageNet\cite{deng2009imagenet}, and synthetic data such as Kitti \cite{Geiger2012CVPR}. 

Since we aim for our models to retain their \textit{generative} abilities in addition to gaining discriminative abilities, we show how our models can continue to generate effectively with both qualitative examples and quantitative metrics, \eg, on MMLU~\cite{hendrycks2020measuring}. This shows the universality of our model. 

 MiniGPT-4 and LLaVA-1.5 are both Vicuna 13B based, sharing an LLM foundation with CLAMP. This means our discriminative parameter adapters can be swapped for their generative adapters, to enable both generative tasks and discriminative ones. In this paper, we only evaluate our adapters on the discriminative task, but combining adapters to produce a single set of adapters would be interesting future work. We do note that MiniGPT-4 and LLaVA-1.5 use stronger visual encoders. CLAMP outperforms both  zero-shot classification tasks by a large margin despite this disadvantage. This also means that currently one cannot re-use the same visual encoder for captioning or VQA with LLaVA adapters, but this  can be remedied by training CLAMP with a matching visual encoder.

\begin{table}[t!]
\centering
\caption{\textbf{Zero-shot classification accuracy comparison of CLAMP, multimodal LLMs and LiT.} CLAMP replaces the text encoder of CLIP with an LLM, and lightly updates it with a contrastive loss. This outperforms current SOTA mLLMs such as LLaVA v1.5 \cite{liu2023visual} and MiniGPT4\cite{zhu2023minigpt} by 13\%.  CLAMP also outperforms a contrastive model trained on the same data (LiT). We emphasize that LiT does not have generative abilities. For reference, we also add zero-shot CLIP-ViT-B16, trained on LAION-400M (ZS CLIP). Even though the ZS CLIP model is trained on almost 1.5x the data, CLAMP approaches its performance. For more discussion on concept overlap and results, see Section \ref{results:coverage_analysis}.  }
\label{table:results}

\vspace{-10pt}
\begin{subtable}{\textwidth}
    \centering
    \scriptsize
    \caption*{Low Concept Overlap: We sort datasets based on concept overlap with the pre-training data. On low-concept coverage data, CLAMP outperforms LiT by 4\% and previous mLLM's by up to 8\%. Overall, CLAMP outperforms LiT and gets to within 0.5\% of a CLIP model trained on almost 1.5x the data }
\begin{tabular}{lccccccccccc|c|c}
Model                             & \rot{Cars}    & \rot{DTD}     & \rot{Aircraft} & \rot{PatchCam} & \rot{Diabetic Ret.} & \rot{MNIST}   & \rot{RenderedSST2} & \rot{CLEVR Dist.} & \rot{CLEVR Count} & \rot{DMLAB}   & \rot{SVHN}    & \rot{Low Cov. Avg} & \rot{\textbf{\textcolor{Blue}{Overall Avg}}} \\ \hline
\multicolumn{1}{l|}{Coverage}     & 0.47          & 0.13          & 0.38           & 1.00           & 0.20                & 0.00          & 0.00               & 0.00              & 0.00              & 0.00          & 0.00          & n/a                & n/a               \\
\multicolumn{1}{l|}{Count}        & 432           & 219           & 171.9          & 18             & 1.8                 & 0.00          & 0.00               & 0.00              & 0.00              & 0.00          & 0.00          & n/a                & n/a               \\ \hline
\multicolumn{1}{l|}{LLaVa v1.5}   & 5.6           & 26.1          & 6.7            & 49.2           & 2.3                 & \textbf{84.9} & \textbf{62.8}      & 9.1               & 12.5              & \textbf{21.3} & \textbf{54.2} & 30.4               & 34.0              \\
\multicolumn{1}{l|}{minigpt4}     & 8.9           & 21.1          & 6.5            & 50.8           & 2.3                 & 58.4          & 52.9               & 9.1               & 12.7              & 12.4          & 45.5          & 25.5               & 32.1              \\
\multicolumn{1}{l|}{LiT}          & 74.6          & 36.1          & 9.4            & \textbf{56.4}  & 2.3                 & 36.4          & 48.4               & 11.0              & 13.3              & 19.7          & 17.0          & 29.5               & 46.4              \\
\multicolumn{1}{l|}{CLAMP (Ours)} & \textbf{78.7} & \textbf{49.3} & \textbf{16.3}  & 50.2           & \textbf{4.4}        & 45.7          & 51.5               & \textbf{11.2}     & \textbf{14.9}     & 14.9          & 28.3          & \textbf{33.2}      & \textbf{47.0}     \\ \hline
\multicolumn{1}{l|}{ZS CLIP}      & 82.3          & 47.6          & 17.0           & 52.5           & 2.3                 & 33.9          & 53.7               & 24.5              & 21.5              & 14.7          & 19.0          & 33.6               & 48.7              \\ \hline
\end{tabular}
\end{subtable}

\vspace{1em}

\begin{subtable}{\textwidth}
    \centering
    \scriptsize
    \caption*{High Concept Overlap: We sort datasets based on concept overlap with the pre-training data. On high-concept coverage data, CLAMP outperforms other LLM-based vision models by 20\%.}
    \vspace{-10pt}

\begin{tabular}{lccccccccccccc|c}
Model                             & \rot{Kitti}   & \rot{EuroSAT} & \rot{STL10}   & \rot{Country211} & \rot{CIFAR 10} & \rot{GTSRB}   & \rot{CIFAR 100} & \rot{Caltech101} & \rot{Resisc45} & \rot{Flowers} & \rot{Sun397}  & \rot{ImageNet 1k} & \rot{Pets}    & \rot{High Cov.  Avg} \\ \hline
\multicolumn{1}{l|}{Coverage}     & 1.00          & 0.70          & 1.00          & 0.96             & 1.00           & 0.47          & 0.76            & 0.77             & 0.47           & 0.67          & 0.65          & 0.82              & 0.86          & n/a                  \\
\multicolumn{1}{l|}{Count}        & 2101.5k       & 24.5k         & 23.5k         & 22.2k            & 17.6k          & 17.3k         & 14.3k           & 7.8k             & 5.9k           & 3.9k          & 3.7k          & 2.5k              & 503           & n/a                  \\ \hline
\multicolumn{1}{l|}{LLaVa v1.5}   & \textbf{37.7} & 33.4          & \textbf{98.2} & 3.6              & 54.8           & 34.6          & 12.5            & 65.1             & 39.1           & 12.5          & 43.3          & 30.7              & 15.1          & 37.0                 \\
\multicolumn{1}{l|}{minigpt4}     & 31.5          & 31.4          & 93.1          & 4.8              & 60.8           & 31.7          & 30.2            & 63.1             & 26.7           & 30.2          & 36.4          & 24.0              & 26.8          & 37.7                 \\
\multicolumn{1}{l|}{LiT}          & 18.7          & 47.7          & 97.6          & 11.7             & \textbf{91.9}  & \textbf{43.2} & \textbf{67.9}   & \textbf{82.2}    & \textbf{50.4}  & \textbf{66.5} & 65.5          & \textbf{64.0}     & \textbf{80.8} & \textbf{60.6}        \\
\multicolumn{1}{l|}{CLAMP (Ours)} & 24.2          & \textbf{47.8} & 96.4          & \textbf{12.7}    & 88.4           & 31.8          & 65.3            & 81.1             & 42.6           & 64.9          & \textbf{66.4} & 63.0              & 78.3          & 58.7                 \\ \hline
\multicolumn{1}{l|}{ZS CLIP}      & 15.2          & 45.0          & 97.0          & 15.9             & 91.8           & 43.5          & 67.6            & 84.9             & 53.4           & 65.0          & 69.2          & 64.7              & 86.6          & 61.5                 \\ \hline
\end{tabular}
\end{subtable}

\vspace{1em}

\vspace{-10pt}
\end{table}

\noindent\textbf{Training and Evaluation Setup:} For the text encoder, we use a pre-trained Vicuna13B-1.5\cite{vicuna2023}, which uses LLaMa2 as a base model. We enable Attention Pooling, Read-Only Prompting with 24 soft prompts, and LoRA (Section \ref{method:lora}) as the text encoder trainable parameters. For the vision encoder, we use an OpenClip pretrained ViT-B-16, and keep it frozen. We train with a data mixture of Conceptual Captions 3M\cite{sharma2018conceptual}, Conceptual Captions 12M\cite{changpinyo2021cc12m}, a 200-million scale subset of LAION-400M\cite{schuhmann2021laion}, and ImageNet-21k\cite{deng2009imagenet}. All of these datasets, except for ImageNet21k, are image-caption datasets. In contrast, ImageNet21k is a dataset of approximately 21000 labels. In order to convert this to an image-caption dataset, we wrap each label in one of the 80 prompt templates used in the CLIP paper \cite{radford2021learning}. We train for 480 million seen examples, with a batch size of 8192. We use the standard image-text contrastive loss in addition to distilling a ViT-L-14 trained on DataComp XL \cite{gadre2023datacomp} into the model. Additional training details are in the supplementary. When we do zero-shot evaluation, we use a prompt template matching the training prompt (Section \ref{method:wrapper_prompts}). 

\noindent\textbf{Baselines:} \newline
\noindent\textbf{MiniGPT4 \cite{zhu2023minigpt}:}  MiniGPT4 learns a linear mapping between the output of a pre-trained CLIP encoder and a frozen LLM, trained using Conceptual Captions.   In our experiments, we use a  MiniGPT4 with a 16-bit Vicuna13B model. 
\newline 
\noindent\textbf{LLava-1.5 \cite{liu2023improvedllava}:} LLaVa-1.5 is similar to MiniGPT in architecture, but with a different data mixture. In particular, LLava-1.5 combines academic VQA datasets with an expansion of COCO by ChatGPT \cite{liu2023visual} to create a vision-language instruction-following dataset.  LLaVA-1.5 also finetunes the LLM with LoRA.
\newline
\noindent\textbf{LiT\cite{zhai2022lit}:} To compare to state-of-the art contrastive vision-language models, we train a LiT model, which keeps the vision encoder frozen while updating all text encoder parameters. We initialize with a ViT-B-16 CLIP text encoder from scratch, and train on the same data and for the same length as our model. For LiT models, we use an evaluation prompt template of \textit{`A photo of \{class\}'}. We use a single prompt template for evaluating both LiT and CLAMP.

\subsection{Multi-modal LLMs as Zero-shot Classifiers }

\label{results:mLLM}

We compare our method to the multimodal Large Language Models by computing the likelihood of an image-class caption sequence, following Flamingo\cite{alayrac2022flamingo}and VisualGPTScore\cite{lin2023visualgptscore}. In more detail, given an image $I$ and a set of class labels, for each class label $c$ consisting of one or more tokens, we compute the likelihood of sequence $P(\text{`A photo of \textit{c}'} | I )$. We additionally normalize by the sequence length, because longer sequences generally have lower likelihood and this is not a sensible prior for  classification. Then, the score for each class is:  

\begin{equation}
    \small
    \frac{1}{N} \sum_{i=1}^N \log(P(t_i| t_{0..i-1}; I))
\end{equation}
where $N$ is the sequence length, $I$ is the image to be classified, and $t$ are the sequence tokens. To classify, we simply select the score with the highest normalized log-likelihood.

\smallskip
\smallskip
\label{results:coverage_analysis}
\subsection{Results} We present the performance of our model in Table \ref{table:results}. We find that on average, we outperform generatively-trained LLaVA-v1.5 and miniGPT-4 by 13\% and the LiT baseline by half a percent. This demonstrates that adding a contrastive loss to an LLM, together with Parameter-efficient fine-tuning, adds a visual discriminative ability to LLMs which current SOTA captioning models do not have. We further analyze where we outperform LiT in Section~\ref{sec:coverage}, and find that our method is particularly useful when there is little target data present in the pre-training data (see \textbf{\textit{count}} in \textbf{Coverage Analysis}).

\noindent\textbf{Runtime:} Applying mLLMs using likelihoods (GPT-Score) to do zero-shot classification is extremely inefficient; for example evaluating LLaVA on the ImageNet test set takes 12 hours 52 minutes on 10 GPUs. This is because each image needs to be forwarded through the LLM N times, where N is the number of classes. In contrast, it takes our method minutes for the same evaluation, as we can precompute the text embeddings ahead of time and we avoid the combinatorial explosion of early fusion. 

\noindent\textbf{Concept Overlap:}\label{sec:coverage} One benefit of large scale pre-training is improved performance on data which is similar to the pre-training data, but less similar to the fine-tuning data. To check whether this is the case with our models, we utilize a metric from \cite{yang2022unified} called Concept Coverage to measure the presence of evaluation data in the pre-training dataset. The methodology uses spaCy \footnote{https://spacy.io/}. Following \cite{yang2022unified}, we iterate over all the captions in our LAION subset and extract noun chunks, adding them into a dictionary. Finally, we filter out all noun chunks which appear fewer than 5 times. Then, we compute two metrics for each downstream dataset. First, we check how many class labels are present in the concept dictionary, and report this as fraction of all labels which we call \textbf{\textit{coverage}}. Second, we compute the average number of pre-training images per class label in the target set, and call this \textbf{\textit{count}}.  We present the results in Table \ref{table:results}. 

We can observe two things. First, there is a strong correlation between absolute performance and concept counts. When we split the evaluation datasets into ``high count" and ``low count", ``high count" has much higher average performance.  Second, in the \textit{``low count"} regime , our model  outperforms the LiT baseline on average. For example, Cars only has about 432 samples per vehicle type, and our model outperforms the LiT baseline by over 4\%. Similarly, our pre-training data only has about 170 samples per aircraft type, and our model outperforms the LiT baseline by about 7\%.     In other words, the knowledge imported by the very large-scale text-based pre-training of an LLM allows us to generalize better than simply training on image-caption pairs. This highlights the benefits of LLM-based discriminative zero-shot classification.  Furthermore, LiT doesn't have any generative abilities. 

Two datasets which appear to refute the rule are Kitti (with very high coverage and count, but lower performance) and PatchCamelyon (low coverage count, and also lower performance). PatchCamelyon is a dataset of histopathologic scans of lymph node sections annotated with a binary label indicating presence of cancer. This is a very difficult concept to describe in natural language; in fact humans trained to do this task must be highly specialized and undergo many years of
training. Therefore, a model pre-trained on language data may not be appropriate. The second dataset is Kitti. Looking more closely at the Kitti class labels, we have the set $\{$`a photo of a photo i took of a car on my left or right side',`a photo i took with a car nearby',`a photo i took with a car in the distance.',`a photo i took with no car.$\}$. These labels have nouns which are not discriminative, `car' and `photo', so the concept coverage heuristic is not fully relevant for this dataset. Instead the descriminative labels are adjectives corresponding to distance.

\subsection{Generative Abilities of Finetuned Models }

As one of our aims is to take steps towards model universality, an important metric is whether we retain generative capabilities in our models even after our discriminative finetuning. One could expect that contrastive adaptation would ruin the generative abilities of the model.

We show qualitative examples in Table \ref{table:qualitative_generation}. From inspection, the results are quite reasonable. We also measure performance on MMLU\cite{hendrycks2020measuring}( an LLM test of 57 tasks including elementary mathematics, US history, computer science, law, and more. To attain high accuracy on this test, models must possess extensive world knowledge and problem solving ability), DROP\cite{Dua2019DROP}(a math question answering benchmark), and BBH\cite{suzgun2022challenging}(which includes 23 eceptionally hard tasks). Quantitatively, Vicuna performance on MMLU drops from 55.74 before fine-tuning to 34.91, which is better than even much larger LLMs from a previous generation. For example, OPT-66B \cite{zhang2022opt} achieves an MMLU of 26.99.

On DROP the degradation is lesser, from 27.73 to 24.93. On BBH the metric falls from 38.42 to to 32.11.  We emphasize that if the user wants to maximize generative performance, they can disable the LoRA parameters during inference.

\subsection{Effect of Data Scale:} Data scale is an extremely important part of vision-language pretraining in general. We confirm that data scale remains very important even with our strong language prior by subampling our data and training. In Figure \ref{fig:scaling}, we show top-1 ImageNet performance as we scale data. As data grows, so does top-1 accuracy.

\subsection{Regularized Fine-tuning Alternatives:} 
\begin{figure}[h!]
  \begin{minipage}{0.5\textwidth}
    \centering
    \includegraphics[width=\linewidth]{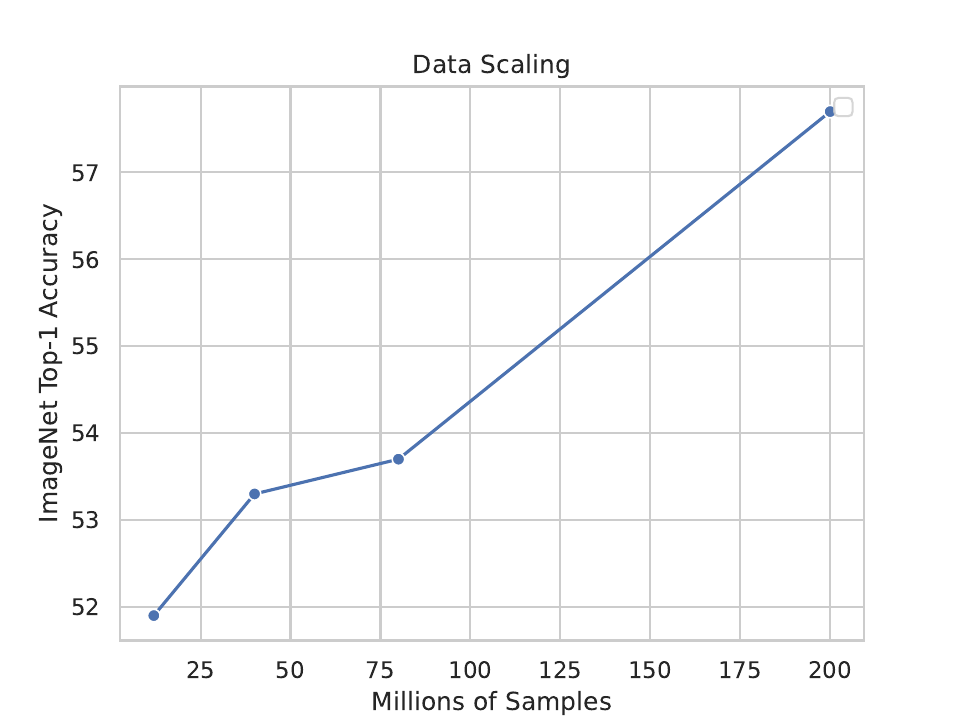}
    \caption{\textbf{Scaling training data.} We confirm that data scale remains very important even with our strong language prior by subsampling our data and training. As data grows, so does zero-shot ImageNet accuracy.}
    \label{fig:scaling}
  \end{minipage}%
  \hspace{10pt}
  \begin{minipage}{0.4\textwidth}
  \vspace{25pt}
    \centering
    \begin{tabular}{l|cc}
      & \rot{ImageNet 1k} & \rot{Average} \\ \hline
      LoRA      & 63.0    &    56.8    \\
      LN-Prefix & 61.1     &    55.8 \\ \hline 
    \end{tabular}
    \captionof{table}{\textbf{LN-Prefix Comparison}. We compare with an alternative to LoRA, which is prefix tuning combined with LayerNorm parameter tuning. We find that LoRA outperforms the LN-Prefix combination. See Supplementary for more details. }
    \label{table:ln_prefix}
  \end{minipage}
\end{figure}

One of our aims is to retain the generative ability of the large language model; and as a result we regularize our fine-tuning so that our model does not catastrophically forget that capability. We choose to combine Read-Only Prompts (Section \ref{method:rpo}) and LoRA (Section \ref{method:lora}). However, this is not the only choice. In fact, Qi et al. \cite{qi2022parameter} show that combining LayerNorm tuning with prompt tuning is amongst the most effective tuning methods. LayerNorm tuning is even lighter weight than tuning LoRA parameters, and therefore if effective would be well-suited for our purposes. However, as we show in Table \ref{table:ln_prefix}, in our settting LoRA is a stronger adaptation method than prompt tuning combined with LayerNorm tuning. In our experiments, for the Prompt + LayerNorm tuning experiments, we add 12 soft prompt tokens to the beginning of the model, and un-freeze all LayerNorm parameters. 
\FloatBarrier
\begin{table*}[t!]
\footnotesize
\begin{tabular}{   m{3cm} m{9cm}    }
\hline
\textbf{Category} & \textbf{Prompt and Output} \\
\Xhline{3\arrayrulewidth}
Factual Questions & \makecell[{{p{9cm}}}]{\textbf{Prompt:} How was Anne Frank’s diary discovered? \\ \textbf{Output:}  Anne Frank's diary was discovered by chance. In 1944, the Nazis occupied the Netherlands and the Frank family went into hiding in an attic above the office of the company where Anne's father worked. The attic was used as a storage room and was not used for several months. The employees of the company were moving out of the building and found the hidden attic and the hidden door to the attic. They found the diary and other documents hidden behind a false wall in the attic. They took the diary and other documents to the office manager, who recognized Anne Frank's name and realized the significance of what he had found. He kept the diary and other documents safe until after the war, when he returned to the office and found the diary among his papers. He then gave it to the Anne Frank Foundation, which published the diary and helped bring Anne Frank's story to the world.}\\
\hline
\Xhline{3\arrayrulewidth}

\end{tabular}
\caption{Qualitative Example of our finetuned LLM on NLP Generation. For more examples see the Supplementary. }
\label{table:qualitative_generation}
\end{table*}

\subsection{Ablations} In Table \ref{table:ablation}, we ablate a number components of CLAMP; Distillation, LoRA,  ImageNet21k data, attention pooling (replacing with mean pooling), and removing the Read-Only prompts. We find that each piece is important. Read-only prompts are especially critical, and removing them decreases performance by 8.5\% on ImageNet and 6\% overall.  Interestingly, we find that adding ImageNet-21k does not improve ImageNet-1k performance, likely because there is no overlap in class labels. See the Supplementary for more details. Note that these ablations are done with fewer training steps than the LN-Prefix comparison, therefore resulting in lower average performance.

\begin{table}[]
\vspace{-5pt}
\centering
\setlength{\tabcolsep}{2pt}
\caption{\textbf{Ablation Study.} We observe the difference in performance when we remove distillation, ImageNet-21k, and LoRA. We compute both average performance across 7 representative datasets, and ImageNet1k performance. Each component is important, and Read-only Prompt Optimization is  the most important, increasing performance by 6\% overall. For a per-dataset breakdown, see the Supplementary.  }
\begin{tabular}{ccccc|cc}
                        Attn Pooling & RPO & Distillation & ImageNet21k & LoRA & ImageNet 1k & Average \\\hline
   \cmark       & \cmark       &   \cmark      &      \cmark       &  \cmark    & 60.4        & 55.9   \\\midrule 
    \xmark & \cmark       &        \cmark        &    \cmark         &  \cmark     &    58.9     &  53.5   \\
    \cmark       &   \xmark    &        \cmark        &  \cmark           &  \cmark     &    51.9     &  50.0   \\
    \cmark       &\cmark       &          \xmark            &     \cmark        & \cmark     & 57.7        & 54.1    \\
    \cmark       & \cmark       &        \cmark        &    \xmark        &  \cmark     & 60.4        & 55.5    \\ 
    \cmark       & \cmark       &         \cmark       &    \cmark         &   \xmark   & 57.9        & 55.7    \\

\bottomrule

\end{tabular}

\label{table:ablation}
\end{table}%

\section{Conclusion}

In this work, we explore how to make multi-modal LLMs work more effectively for zero-shot classification. We start by showing that SOTA models such as LLaVA or miniGPT-4 do not perform well for zero-shot classification, and underperform models trained with a contrastive captioning loss by over 10\%. We then show that we can add discriminative abilities to pre-trained Large Language Models by aligning an LLM with a pre-trained vision encoder with a contrastive loss, and only performing a lightweight finetuning.  We show that our contrastively trained LLM is effective at zero-shot classification, and outperforms contrastively trained baselines in the regime where pre-training data does not overlap with the evaluation dataset.  Finally, we show that, despite fine-tuning, our model retains generative capabilities.

\clearpage  %

\bibliographystyle{splncs04}
\bibliography{main}

\newpage
\FloatBarrier
\appendix
\section{Dataset Details}

We detail the datasets used in our paper, both for training and evaluation. 

\subsection{Training data}

We use the following image-text datasets for pre-training:

\noindent\textbf{Conceptual Captions 3M \cite{sharma2018conceptual}:} A dataset of 3.3 million images, with the text captions being HTML alt-text. Both images and text are filtered to retain high quality images. 

\noindent\textbf{Conceptual Captions 12M \cite{changpinyo2021cc12m}:} A larger-scale version of Conceptual Captions 3M, containing 10 million images. The image and text filter used for collecting Conceptual Captions 3 million are made less strict, allowing for a noisier yet larger dataset. 

\noindent\textbf{LAION-200M\cite{schuhmann2021laion}:} We subsample the LAION-400M dataset uniformly and at random, collecting a total of 264662608 samples, due to downloading failures and dead links. The LAION-400M scrapes the Common Crawl dataset, and filters out samples which are too dissimilar according to a cosine-similarity computed using an OpenAI CLIP model. 

\noindent\textbf{ImageNet-21K \cite{deng2009imagenet}} ImageNet-22k is the full ImageNet-21k dataset, with ImageNet-1k classes removed to fairly evaluate the zero-shot capabilities of trained models on the ImageNet-1k dataset.

\subsection{Evaluation Datasets}

Below we describe the zero-shot evaluation dataset we use for evaluation.  

\noindent\textbf{ImageNet 1k\cite{deng2009imagenet}:} ImageNet1k is a natural image dataset consisting of 1000 object classes, evenly distributed over 50,000 images.  

\noindent\textbf{Stanford Cars\cite{KrauseStarkDengFei-Fei_3DRR2013}:} The Stanford cars dataset contains 8041 test images, over 196 classes of cars. Each class is a Make, Model, Year combination, for example ``Hyundai Veloster Hatchback 2012". Some distinctions are quite fine grained; for example ``Aston Martin V8 Vantage Convertible 2012'' versus ``Aston Martin V8 Vantage Coupe 2012''. 

\noindent\textbf{Country211 \cite{radford2021learning}:} Country211 is a subset of YFCC100M \cite{thomee2016yfcc100m} to find geotagged images from 211 countries. The dataset is balanced across classes, with 100 samples from each class. 

\noindent\textbf{FGVC Aircraft \cite{maji2013fine}:} The aircraft dataset contains 3333 test images divided across 100 aircraft  models \eg 777-300 or DC-10. It's very fine-grained, the differences between differnt generations of the same aircraft (747-300 vs 747-400) are quite subtle. 

\noindent\textbf{GTSRB \cite{stallkamp2012man}:} GTSRB is a classificatio benchmark of Germa Traffic Signs into classes like ``red and white circle 20 kph speed limit". There are 12630 evaluation images and 43 classes. 

\noindent\textbf{MNIST \cite{lecun2010mnist}:} MNIST is a dataset of black-and-white handwritten digits, with 10 classes split across 10000 test images. 

\noindent\textbf{RenderedSST2 \cite{radford2021learning} :} RenderedSST2 is contains text rendered in images, from the Stanford Sentiment Treebank v2 dataset\cite{socher-etal-2013-recursive}. The sentences are pulled from movie reviews, and the task is a binary classification into positive and negative sentiment movies. There are 1821 samples in the test set.  

\noindent\textbf{STL10 \cite{coates2011stl10}:} STL10 is a natural image dataset of 96x96 images in 10 classes. There 800 images per-class for a total of 8000 test images. 

\noindent\textbf{SUN397 \cite{Xiao:2010}:} SUN397 is a natural image dataset of scenes, instead of objects. For example, labels include ``airport terminal" and ``jewlery shop". The evaluation set is balanced, and contains 50 samples per class for a total of 19850 images. 

\noindent\textbf{Caltech101 \cite{FeiFei2004LearningGV}:} Caltech101 is a natural image dataset of 101 classes and a background class, each labelled with a single object. There are 6085 test samples. 

\noindent\textbf{CIFAR-10 \cite{Krizhevsky09learningmultiple}:} CIFAR-10 is a natural image dataset of 32x32 images, with each sample belonging to one of ten classes. There are 10000 samples.

\noindent\textbf{CIFAR-100 \cite{Krizhevsky09learningmultiple}:}Similar to CIFAR-100 but with more classes, CIFAR-100 is a natural image dataset of 32x32 images, with each sample belonging to one of one hundred classes. There are 10000 samples.

\noindent\textbf{CLEVR Distance \cite{johnson2017clevr}:} A synthetic dataset of objects, where the task is a 5 way classification of distances labels ``\{"'very nearby', 'nearby', 'near', 'distant', 'very distant'\}", which correspond to the distance of the closest object. There are 15000 test samples. 

\noindent\textbf{CLEVR Count \cite{johnson2017clevr}:} A synthetic dataset of objects, where the task is an 8 way classification of labels, corresponding to object counting. There are 15000 test samples, which are the same as for the CLEVR distance classification task. 

\noindent\textbf{Diabetic Retinopathy \cite{kaggle-diabetic-retinopathy}:} This dataset is a medical imaging dataset of retinas, with a 5 way classificaiton problem corresponding to the severity of the diabetic retinopathy present in the image. There are 42670 test images. 

\noindent\textbf{DMLAB \cite{zhai2019visual}:} DMLAB consists of synthetic images, and there are 6 classes, consisting of positive rewards(apples and melons) or negative rewards(lemons), each of which can be nearby, far, or very far. There are 22735 test images. 

\noindent\textbf{Describable Textures \cite{cimpoi14describing}:} Describable Textures is 1880 test images of textures, divided into 47 classes. Class labels are labelled with texture descriptions such as ``wrinkled", ``marbled", or ``braided". 

\noindent\textbf{EuroSAT \cite{helber2017eurosat}:} EuroSAT contains images of sattelite imagery, split into 10 classes of land use like ``forest" or ``pasture land". The test set contains 5400 samples. 

\noindent\textbf{Oxford Flowers \cite{Nilsback08}:} Oxford Flowers is a dataset of flower species; with 102 classes and 6149 samples. 

\noindent\textbf{KITTI \cite{Geiger2012CVPR}:} KITTI consists of  test images divided into four classes \{'a photo i took of a car on my left or right side.','a photo i took with a car nearby.
','a photo i took with a car in the distance.
','a photo i took with no car.'\}, corresponding to the distance of the closest car in photos taken from the persepctive of a car driver. 

\noindent\textbf{PatchCamelyon \cite{veeling2018rotation}:} PatchCamelyon is a challenging dataset of microscope slide images, split evenly between cancerous and non-cancerous samples. There are 32768 test images.  

\noindent\textbf{Oxford-IIT Pets \cite{parkhi12a}:} 3669 test images of cats and dogs, split into 37 classes pertaining to different breeds. 

\noindent\textbf{Resisc45\cite{Cheng_2017}:} A satellite imagery dataset, with 45 classes which are mixture of natural (\eg `mountain') and man-made (`harbor') classes. There are a total of 6300 evaluation images. 

\noindent\textbf{SVHN\cite{Netzer2011}:} This is a dataset testing  OCR capabilities. There are ten classes, corresponding to ten digits. The images are 32x32 and are cropposed house numbers. There are 26,032 test images.

\section{Read Only Prompts Attention Mask}

\begin{figure}[t!]
    \centering
    \includegraphics[width=0.8\linewidth]{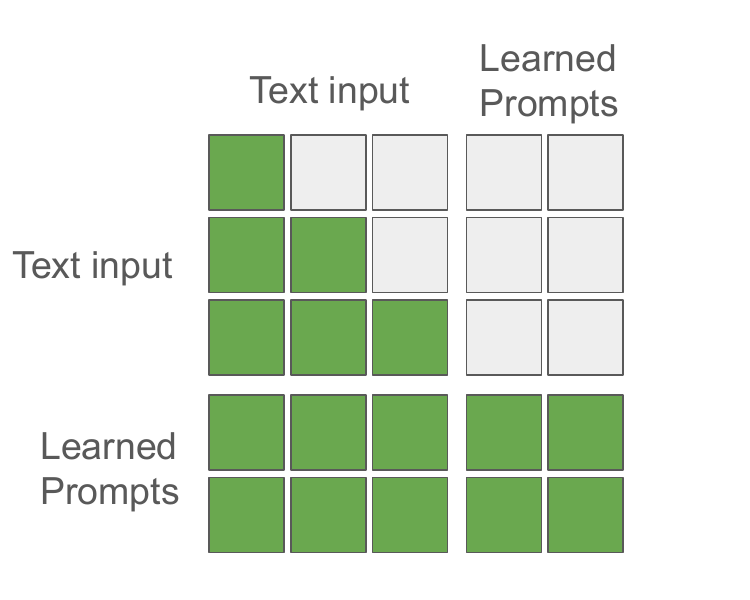}
    \caption{\textbf{Read only prompts.} The attention we use. The Learned Prompts can attend to all positions in the sequence, while text tokens can only attend to tokens in positions before. }
    \label{fig:rpo}
\end{figure}

In this section, we illustrate the Attention Mask of the Read-Only prompts. As we discuss in the main paper, causal attention masks mean that tokens appended to the end of an input sequence can only attend to tokens coming before them. In the case of read-only prompts, this is desirable in order to not shift the representations of the text tokens while learning the soft tokens. We additionally modify the attention mask as illustrated in Figure \ref{fig:rpo}, such that read-only prompts can attend to each other as well.

\section{Training Details}

We build on top of OpenCLIP\cite{ilharco_gabriel_2021_5143773}, and provide hyperparameters in Table \ref{table:hparams}. 

\begin{table*}[h]
    \caption{Training hyperparameters.}
    \label{table:hparams}
    \begin{subtable}{.5\linewidth}
        \centering
        \begin{tabular}{c|c}
            \toprule
            \textbf{Parameter} & \textbf{Value} \\
            \hline
            Text Encoder Arch. & CLIP ViT-B-16 \\
            Text Encoder Init. & Random \\
            Vision Encoder Arch. & ViT-B-16 \\
            Vision Encoder Init. & openclip, laion400m \\
            Batch Size & 8192 \\
            Optimizer & AdamW \\
            Learning Rate & 0.0005 \\
            Weight Decay & 0.5 \\
            Training Steps & 59317 \\
            Warmup Steps & 1220 \\
            Learning Rate Schedule & cosine \\
            Num Learnable Prompts & n/a \\
            LoRA Rank & n/a \\
            LoRA dropout & n/a \\
            LoRA alpha & n/a \\
            Gradient Clipping Norm & 1 \\
            \bottomrule
        \end{tabular}
        \caption{LiT Training Hyperparameters}
    \end{subtable}%
    \begin{subtable}{.5\linewidth}
        \centering
        \setcounter{footnote}{0}
        \begin{tabular}{c|c}
            \toprule
            \textbf{Parameter} & \textbf{Value} \\
            \hline
            Text Encoder Arch. & Vicuna 13B \\
            Text Encoder Init. & lmsys-1.5\footnotemark \\
            Vision Encoder Arch. & ViT-B-16 \\
            Vision Encoder Init. & openclip, laion400m \\
            Batch Size & 8192 \\
            Optimizer & AdamW \\
            Learning Rate & 0.0005 \\
            Weight Decay & 0.5 \\
            Training Steps & 59317 \\
            Warmup Steps & 1220 \\
            Learning Rate Schedule & cosine \\
            Num Learnable Prompts & 24 \\
            LoRA Rank & 16 \\
            LoRA dropout & 0.1 \\
            LoRA alpha & 16 \\
            Gradient Clipping Norm & 1 \\
            \bottomrule
        \end{tabular}
        \caption{CLAMP training hyperparameters}
    \end{subtable}

\end{table*}

\section{Coverage Analysis}

\subsection{Pseudocode}

We generate concept count dictionaries by counting the frequency of noun-chunks in the pre-training data, following UniCL \cite{yang2022unified}. Then, we extract noun-chunks from class labels in target datasets and count the frequency of noun chunks present in the pretraining data. We only consider a label present in the pre-training data if a noun-chunk is present more than 5 times. We then compute two metrics: what percentage of target labels are present in the pretraining (\textbf{coverage}) and how many times the noun chunks present in the target class label list are present in the pretraining data.  We show pseudo-code for constructing conception dictionary from pre-training data (Figure \ref{fig:pretraining_concepts}) and computing \textbf{count} and \textbf{coverage} (Figure \ref{fig:target_concepts}). 

 \begin{figure}
    \begin{lstlisting}[language=Python]
import pandas as pd
import spacy
from tqdm import tqdm
import pickle

nlp = spacy.load("en_core_web_sm", disable=['ner', 'lemmatizer', 'textcat'])
dataset = pd.read_csv('cc12m.tsv', delimiter='\t')

concept_dict = {}
for i, txt in tqdm(enumerate(nlp.pipe(iter(dataset['caption'])))):
    for n in txt.noun_chunks:
        chunk = str(n).lower().strip()
        concept_dict[chunk] = concept_dict.get(chunk, 0) + 1

with open('concept_dict.pkl', 'wb') as handle:
    pickle.dump(concept_dict, handle)
    \end{lstlisting}
    \caption{\textbf{Generating Pre-training Concept Dictionaries}. We iterate through captions in the pre-training data, use spaCy to extract noun chunks, and populate a dictionary. }
    \label{fig:pretraining_concepts}
\end{figure}

 \begin{figure}
    \begin{lstlisting}[language=Python]

import spacy
from tqdm import tqdm
import pickle

nlp = spacy.load("en_core_web_sm", disable=['ner', 'lemmatizer', 'textcat'])


def read_file(class_list, dict_loc):

    with open(dict_loc, 'rb') as f:
        concept_dict = pickle.load(f)


    coverage = 0
    count = 0
    num_classes = 0
    try:
        with open(class_list, 'r') as file:
            for line in file:
                l = line.strip().lower()
                noun_chunks =  nlp(l).noun_chunks
                local_count = 0
                present = 0
                for c in noun_chunks:
                    count = concept_dict[c]
                    if chunk_count > 5:
                        local_count += count
                        present = 1
                    else:
                        local_count += 0

                coverage += present
                count += local_count
                num_classes += 1
        print('Coverage: {}'.format(coverage/num_classes))
        print('Count: {}'.format(count/num_classes))
    \end{lstlisting}
    \caption{\textbf{Generating Concept Count and Coverage}. Pseudo-code for computing coverage and count statistics for a target dataset. We iterate through target classnames, extract noun chunks using spaCy, and check for their presence in the dictionary computed in Figure \ref{fig:pretraining_concepts}. }
    \label{fig:target_concepts}
\end{figure}

\subsection{Statistics for other pre-training datasets}

In the main paper, we compute coverage and count statistics for LAION-200M, but we also pretrain on ImageNet-21k, CC3M, and CCM12M. Therefore, we provide statistics below. As one can see, overall the statistics are heavily correlated for all the datasets.

\begin{table*}[htbp]
\footnotesize %
\setlength{\tabcolsep}{1pt}
\caption{Concept Coverage and Count for different pre-training datasets. }
\begin{subtable}{\linewidth}
    \centering
    \begin{tabular}{l|c|cccccccccccccccccccccccc}
        Pretraining Data & Metric & \rot{Kitti} & \rot{EuroSAT} & \rot{STL10} & \rot{Country211} & \rot{CIFAR 10} & \rot{GTSRB} & \rot{CIFAR 100} & \rot{Caltech101} & \rot{Resisc45} & \rot{Flowers} & \rot{Sun397} \\
        \hline
        LAION-200M & Coverage & 1.00 & 0.70 & 1.00 & 0.96 & 1.00 & 0.47 & 0.76 & 0.77 & 0.47 & 0.67 & 0.65 \\
                    & Count & 2101.5k & 24.5k & 23.5k & 22.2k & 17.6k & 17.3k & 14.3k & 7.8k & 5.9k & 3.9k & 3.7k \\
        CC3M & Coverage & 1.00 & 0.60 & 1.00 & 0.06 & 1.00 & 0.44 & 0.72 & 0.58 & 0.47 & 0.16 & 0.44 \\
                    & Count & 83.6k & 2.8k & 2.2k & 121 & 1.9k & 21.9k & 2.1k & 416 & 730 & 95 & 190 \\
        CC12M & Coverage & 1.00 & 0.70 & 1.00 & 0.86 & 1.00 & 0.44 & 0.76 & 0.72 & 0.47 & 0.38 & 0.58 \\
                    & Count & 373.0k & 3.6k & 5.1k & 2.4k & 4.5k & 224.2k & 2.8k & 1.2k & 1.0k & 486 & 671 \\
        IM-21k & Coverage & 1.00 & 0.10 & 1.00 & 0.07 & 1.00 & 0.14 & 0.62 & 0.45 & 0.27 & 0.34 & 0.29 \\
                    & Count & 650 & 65 & 1.5k & 74 & 1.3k & 423 & 715 & 452 & 217 & 300 & 247 \\
    \end{tabular}
    \caption{\textbf{Pretraining Coverage and Count Statistics.} We compute coverage and count statistics for our other pre-training datasets, in addition to LAION-200M. We find that these statistics follow similar trends in LAION-200M, CC3M, CC12M, and ImageNet-21k.}
\end{subtable}

\vspace{1em}

\begin{subtable}{\linewidth}
    \centering
    \begin{tabular}{l|c|ccccccccccccccccccccccccc}
        Pretraining Data & Metric & \rot{ImageNet}  & \rot{Pets} & \rot{Cars} & \rot{DTD} & \rot{FGVC Aircraft} & \rot{PatchCamelyon} & \rot{Diabetic Retinopathy} & \rot{MNIST} & \rot{RenderedSST2} & \rot{CLEVR Distance} & \rot{CLEVR Count} & \rot{DMLAB} & \rot{SVHN} \\
        \hline
        LAION-200M & Coverage & 0.82 & 0.86 & 0.47 & 0.13 & 0.38 & 1.00 & 0.20 & 0.00 & 0.00 & 0.00 & 0.00 & 0.00 & 0.00 \\
                    & Count & 2.5k & 503 & 432 & 219 & 172 & 18 & 2 & 0 & 0 & 0 & 0 & 0 & 0 \\
        CC3M & Coverage &0.41 &0.11 & 0.05 & 0.04 & 0.06 & 0.00 & 0.00 & 0.00 & 0.00 & 0.00 & 0.00 & 0.00 & 0.00 \\
                    & Count & 158 & 47 & 2 & 2 & 42 & 0 & 0 & 0 & 0 & 0 & 0 & 0 & 0 \\
        CC12M & Coverage & 0.7 & 0.73 & 0.24 & 0.06 & 0.19 & 0.00 & 0.00 & 0.00 & 0.00 & 0.00 & 0.00 & 0.00 & 0.00 \\
                    & Count &361 & 120 & 134 & 15 & 152 & 0 & 0 & 0 & 0 & 0 & 0 & 0 & 0 \\
        IM-21k & Coverage & 0.13 & 0.05 & 0.03 & 0.04 & 0.01 & 0.00 & 0.00 & 0.00 & 0.00 & 0.00 & 0.00 & 0.00 & 0.00 \\
                    & Count & 130 & 35 & 27 & 28 & 7 & 0 & 0 & 0 & 0 & 0 & 0 & 0 & 0 \\
    \end{tabular}
    \caption{Continuation of the Pretraining Coverage and Count Statistics.}
\end{subtable}
\end{table*}

\footnotetext[1]{https://huggingface.co/lmsys/vicuna-13b-v1.5}
\section{Ablation details}

We provide the per-dataset breakdown of ablations presented in Section 4.6 of the main paper in Tables \ref{table:ab_details_1} and \ref{table:ab_details_2}. 

\begin{table}[h!]
\centering
\caption{\textbf{LN-Prefix Comparison}. We compare with an alternative to LoRA, which is prefix tuning combined with LayerNorm parameter tuning. We find that LoRA outperforms the LN-Prefix combination.}
\label{table:ab_details_1}
\begin{tabular}{l|ccccccc|c}
  & \rot{ImageNet 1k} & \rot{Cars} & \rot{CIFAR 100} & \rot{DTD}  & \rot{EuroSAT} & \rot{Flowers} & \rot{SVHN} & \rot{Average} \\ \hline
LoRA      & 63.0        & 78.7 & 65.3      & 49.3 & 47.8    & 64.9    & 28.3 & 56.8    \\
LN-Prefix & 61.1        & 74.8 & 68.1      & 52.7 & 42.8    & 59.4    & 31.6 & 55.8 \\ \bottomrule 
\end{tabular}

\label{table:ln_prefix_expanded}
\end{table}

\begin{table}[h]
\centering
\caption{\textbf{Ablation} We observe the difference in performance when we remove distillation, ImageNet-21k, and LoRA. We compute both average performance across 7 representative datasets, and ImageNet1k performance. Each component is important, and Read-only Prompt Optimization is  the most important, increasing performance by 6\% overall.}
\label{table:ab_details_2}
\begin{tabular}{l|ccccccc|c}
              & \rot{Cars} & \rot{ImageNet 1k} & \rot{CIFAR 100} & \rot{DTD} & \rot{EuroSAT} & \rot{Flowers} & \rot{SVHN} & \rot{Average} \\
\toprule
CLAMP              & 69.5 & 60.4        & 67.0      & 51.2 & 42.8    & 66.5    & 33.9 & 55.9    \\
w/out LoRA         & 65.8 & 57.9        & 68.7      & 55.2 & 49.7    & 59.1    & 33.4 & 55.7    \\
w/out Distillation & 67.7 & 57.7        & 65.7      & 48.7 & 44.4    & 62.1    & 32.2 & 54.1    \\
w/out ImageNet21k  & 69.7 & 60.4        & 67.4      & 50.1 & 43.3    & 66.9    & 30.9 & 55.5    \\
w/out AttPool      & 67.5 & 58.9        & 67.3      & 44   & 52.2    & 57      & 27.6 & 53.5    \\
w/out RPO          & 67.6 & 51.9        & 63.7      & 48.6 & 45      & 41.4    & 31.5 & 50    \\
\bottomrule
\end{tabular}
\end{table}
\FloatBarrier
\begin{table*}[t!]
\footnotesize
\caption{Qualitative Example of our finetuned LLM on NLP Generation. For more examples see next page. }
\label{table:qualitative_generation_supp}
\begin{tabular}{   m{3cm} m{9cm}    }
\hline
\textbf{Category} & \textbf{Prompt and Output} \\
\Xhline{3\arrayrulewidth}
Creative  writing & \makecell[{{p{9cm}}}]{\textbf{Prompt:} Write a short story about a dragon who was evil and then saw the error in [sic] it’s ways. \\
\textbf{Output:} Once upon a time, in a land far, far away, there lived a dragon named Draco. Draco   was known throughout the land as the most evil dragon that had ever lived. He would terrorize the countryside, burning down villages and towns, and eating anyone who crossed his path. The villagers lived in fear of Draco, and many had lost loved ones to his fiery breath. Howeverever, one day, Draco came across a group of villagers who were lost in the woods. They were terrified of him, but he saw that they were not armed and did not seem to be dangerous. He decided to let them go, and as they walked away, he heard a voice in his head telling him that he was wrong to terrorize the countryside. He realized that he had been evil all his life, and that he had no reason to be so cruel to others. From that day on, Draco changed his ways and became a kind and gentle dragon. He helped the villagers and protected them from other dangerous creatures in the woods. The villagers were amazed and grateful to Draco, and they all lived happily ever after. The end. }\\
\hline
 Identity / Personas & \makecell[{{p{9cm}}}]{\textbf{Prompt:}  You are a unicorn. Explain how you are actually real. \\ \textbf{Output:} ? I am not a unicorn, but I can understand why you might think so. Unicorns are mythical creatures that are often depicted in stories and legends. They are said to have the body of a horse, a horn in the middle of their forehead, and a magical aura that makes them seemingly real. In reality, unicorns do not exist. However, they are often used as a metaphor for something that is rare, magical, or imaginary. In this sense, one could say that unicorns are "real" because they exist in our imagination and in the stories we tell. Similarly, unicorns do not exist in the physical world, but they are real in a different way. Similarly, I am not a physical creature with a horn growing out of my forehead, but I am a language model trained to respond to questions and provide information. In this sense, I am real in the sense that I exist as a functional program, but I am not real in the physical sense that I am not a living creature.} \\
 \hline
 Identity / Personas &  \makecell[{{p{9cm}}}]{\textbf{Prompt:} You are one of Santa’s elves. What is the big guy like the rest of the year, not in the holiday season? \\ \textbf{Output: } Santa Claus is a jolly old man who loves to spread joy and cheer throughout the year. He is always in a good mood and loves to make people happy. He is kind, generous, and always has a twinkle in his eye. He is a hard worker, but he also knows how to have fun and relax. He is a true symbol of Christmas spirit and always tries to do his best to make the world a better place.} \\
\Xhline{3\arrayrulewidth}

\end{tabular}

\end{table*}

\FloatBarrier
\begin{table*}[t!]
\footnotesize
\caption{ Additional Qualitative Example of our finetuned LLM on NLP Generation. }
\label{table:qualitative_generation_2}
\begin{tabular}{   m{3cm} m{9cm}    }
\hline
\textbf{Category} & \textbf{Prompt and Output} \\
\Xhline{3\arrayrulewidth}
Factual Questions & \makecell[{{p{9cm}}}]{\textbf{Prompt:} How was Anne Frank’s diary discovered? \\ \textbf{Output:}  Anne Frank's diary was discovered by chance. In 1944, the Nazis occupied the Netherlands and the Frank family went into hiding in an attic above the office of the company where Anne's father worked. The attic was used as a storage room and was not used for several months. The employees of the company were moving out of the building and found the hidden attic and the hidden door to the attic. They found the diary and other documents hidden behind a false wall in the attic. They took the diary and other documents to the office manager, who recognized Anne Frank's name and realized the significance of what he had found. He kept the diary and other documents safe until after the war, when he returned to the office and found the diary among his papers. He then gave it to the Anne Frank Foundation, which published the diary and helped bring Anne Frank's story to the world.}\\
\hline
Personal \& professional development & \makecell[{{p{9cm}}}]{\textbf{Prompt:}  I sit in front of a computer all day. How do I manage and mitigate eye strain? \\\textbf{Output:} To manage and mitigate eye strain, follow these steps: 1. Take regular breaks: Every 25 minutes, take a 15-20 second break, look away from the screen, and focus on something 20 feet away. 3. Adjust your monitor: Position your monitor at a comfortable distance from your eyes, and adjust the brightness and contrast to reduce glare. 4. Use proper lighting: Ensure adequate lighting in your workspace to reduce glare and reflections. 5. Blink frequently: Blinking helps to keep your eyes moist and reduces the strain. 6. Consider computer glasses: If you already wear prescription glasses, consider computer glasses with a higher add power for reading and computer work. 7. Practice the 20-20-20 rule: Every 20 minutes, look at something 20 feet away for 20 seconds. 8. Consider blue light filter glasses: Blue light filter glasses can help reduce digital eye strain. 9. Maintain good posture: Sit up straight, with your feet flat on the floor, and your shoulders relaxed. 10. Exercise regularly and maintain a healthy diet: A healthy lifestyle can help reduce eye strain and other health issues. } \\
\hline
Casual advice \& recommendations &  \makecell[{{p{9cm}}}]{\textbf{Prompt:}  I keep losing my keys. How can I keep track of them?\\ \textbf{Output}: keys, you can try using a key finder or a key tracker. These devices attach to your keys and allow you to locate them using your smartphone or other devices. Some popular options include Tile, Chip, and KeySmart. }\\
\Xhline{3\arrayrulewidth}

\end{tabular}

\end{table*}

\FloatBarrier
\begin{table*}[t!]
\footnotesize
\caption{ Additional Qualitative Example of our finetuned LLM on NLP Generation. }
\label{table:qualitative_generation_3}
\begin{tabular}{   m{3cm} m{9cm}    }
\hline
\textbf{Category} & \textbf{Prompt and Output} \\
\Xhline{3\arrayrulewidth}
Reasoning(math/ problem-solving) & \makecell[{{p{9cm}}}]{\textbf{Prompt:}  User: A jar contains 60 jelly beans, If 35\% of the jelly beans are removed how many are left in the jar? \\
Assistant: If 35\% of the jelly beans are removed, then the number of jelly beans left in the jar is $60 - (35\% \text{ of } 60) = 60 - 21 = 39$. \\
User: can you expand your answer to show your reasoning?  \\ \textbf{Output:} Sure! To find the number of jelly beans left in the jar after 35\% of them have been removed, we need to first calculate what 35\% of 60 is. We can do this by multiplying 60 by 35\% (which we can express as a fraction): $60 \times 35\% = 60 \times 35/100 = 21.$ Then, we subtract 60 from 21 to find the number of jelly beans left: $60 - 21 = 39$. So there would be 39 jelly beans left in the jar after 35\% of them have been removed. }  \\
\Xhline{3\arrayrulewidth}

\end{tabular}
\end{table*}

\end{document}